\documentclass[journal,transmag]{IEEEtran}
\usepackage{amssymb}
\usepackage{upgreek}
\usepackage{multirow}
\usepackage{multicol}
\usepackage[T1]{fontenc}

\usepackage{cite}

\ifCLASSINFOpdf
\usepackage[pdftex]{graphicx}
\else
\fi
\usepackage{amsmath}
\usepackage{algorithmic}
\begin{document}
%
\title{Learning Deep Representations by Mutual Information for Person Re-identification
}


\author{\IEEEauthorblockN{Peng Chen\IEEEauthorrefmark{1},
Tong Jia\IEEEauthorrefmark{1},
Pengfei Wu\IEEEauthorrefmark{1}, 
Jianjun Wu\IEEEauthorrefmark{1}, and
Dongyue Chen\IEEEauthorrefmark{1}
}
\IEEEauthorblockA{\IEEEauthorrefmark{1}College of Information Science and Engineering,
Northeastern University, Shenyang, China}
\thanks{Dongyue Chen is the corresponding author}}

%



\IEEEtitleabstractindextext{%
\begin{abstract}
Most existing person re-identification (ReID) methods have good feature representations to distinguish pedestrians with deep convolutional neural network (CNN) and metric learning methods. However, these works concentrate on the similarity between encoder output and ground-truth, ignoring the correlation between input and encoder output, which affects the performance of identifying different pedestrians. To address this limitation, We design a Deep InfoMax (DIM) network to maximize the mutual information (MI) between the input image and encoder output, which doesn't need any auxiliary labels.
To evaluate the effectiveness of the DIM network, we propose end-to-end Global-DIM and Local-DIM models. Additionally, the DIM network provides a new solution for cross-dataset unsupervised ReID issue as it needs no extra labels. The experiments prove the superiority of MI theory on the ReID issue, which achieves the state-of-the-art results.
\end{abstract}

}

\maketitle

\IEEEdisplaynontitleabstractindextext

%
\IEEEpeerreviewmaketitle


\section{Introduction}
%
%
%
%
\IEEEPARstart{R}{eID} 
\label{section:1}
is a branch of image retrieval, which aims to search a query pedestrian image from gallery set. Although the ReID issue achieves dramatic progress with the development of CNN, it is still a challenging problem due to the complexity of pedestrian. The feature representations obtained by the CNN extractor cannot represent the input pedestrian exactly because pedestrians themselves have great similarities in the classification task.

\begin{figure}
    \centering
    \includegraphics[width=3.4in]{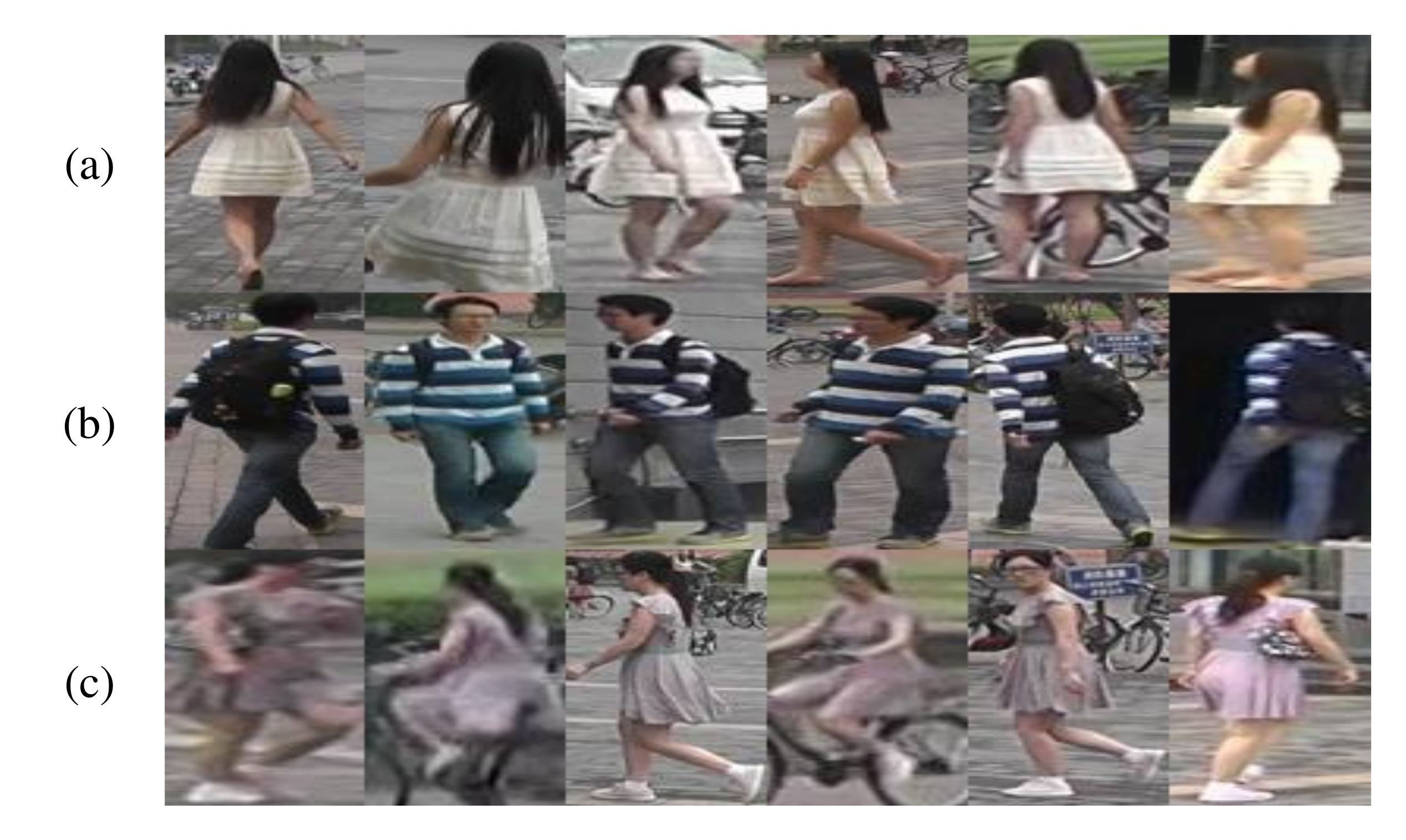}
    \caption{Illustration of the variance in the same pedestrian.  (a), (b), (c) represent different pedestrians.}
    \label{fig:my_label1}
\end{figure}

Related studies are performed on this problem. For pedestrian similarity issue, \cite{Zheng2016,Yao2017} adopted a deep convolutional network and body parts to learn discriminative representations. They rely on powerful CNN to extract feature representations. \cite{Zheng2016},\cite{Varior2016},\cite{Ding2015} utilized metric loss to train a deep model for discriminative representations. However, these works concentrate on the encoder output and ground-truth, neglecting the correlation between the input and encoder output. Specifically, for different images of the same pedestrian, the above methods extract the common features of the pedestrian, ignoring the unique characteristics and details of different images, which may damage the generalization of the model. For examples, in Fig \ref{fig:my_label1}, (a), (b) and (c) are the three different pedestrians of which have six different images respectively. The common features of the pedestrian cannot describe the details clearly, which hinder the capability of feature representations.

In this paper, we focus on the correlation between input and encoder output to enhance feature representations for each image. We design a DIM network to maximize the MI between the input pedestrian and encoder output. Based on the DIM network, We propose an end-to-end Global-DIM network to capture the discriminative feature representations. Furthermore, inspired by the success of the part-based ReID strategy in \cite{Sun2017}, we implement an end-to-end Local-DIM network to maximize the MI between the parts of pedestrians. 

Besides, the feature representations are also limited to the dataset. Because the target dataset in the real surveillance video has different feature distribution and is not labeled, the model trained on the source dataset is difficult to apply. To address such problem, various unsupervised domain adaptation methods try to fit the distribution of source and target dataset \cite{Fan2017},\cite{Deng2017},\cite{Zhong2017a}. However, these methods usually rely on auxiliary information like camera labels and are time-consuming in the training phase. Our DIM network provides a feasible solution for unsupervised ReID issue, as it has no auxiliary cues and is easy to train. We propose a Transfer Deep InfoMax (TF-DIM) model to extract the specific characteristics of pedestrians in the target dataset for unsupervised ReID.

Our contributions can be summarized into three aspects: 1). We propose a DIM network to maximize the MI between the input image and encoder output, therefore boosting the capability of feature representations. 2). Based on the IDE and PCB baseline, we build Global-DIM and Local-DIM networks to verify the effectiveness of DIM network in ReID issue. 3). Based on the unlabeled DIM loss, We construct a TF-DIM model to obtain feature representations with a well-generalized performance for unsupervised ReID issue. 4). By using the theory of MI, we propose an end-to-end high-quality ReID model for both supervised and unsupervised ReID issues, which does not need any extra knowledge cues.

\section{Related works}

\subsection{Supervised-learned ReID}
In order to deal with the problem of pedestrian similarities, a wide range of methods are proposed to achieve discriminative feature representations. They can be classified by feature and metric learning. For feature learning methods, considering the characteristics of the pedestrian, one idea is parsing pedestrian into part features. Wei et al. \cite{Wei2017a} proposed a Global-Local-Alignment Descriptor (GLAD) to partition the pedestrian. Su et al. \cite{Su2017} utilized pedestrian part cues to boost feature representations. Additionally, the attention mechanism \cite{Xu2015} is widely employed in deep ReID. Liu et al. \cite{Liu2017} and Zhao et al. \cite{Zhao2017} enhanced the feature representations with attention mechanism, in which the network can choose the important part. \cite{Sun2017} proposed an end-to-end attention ReID network for robust feature representations. For metric learning methods, \cite{Varior2016} computes the contrastive loss to distinguish similar pedestrians. \cite{Ding2015} applied triplet loss for better feature representations. However, these methods neglect the correlation between input and encoder output, which may damage the generalization of the model. In this paper, we introduce the MI theory \cite{Hjelm2018} to improve the capacity of feature representations.

\subsection{Unsupervised-learned ReID}
The big size of unlabeled target dataset in real surveillance scenarios is difficult to recognize in the pre-trained model based on source labeled dataset. Xiang et al. \cite{Kodirov} utilized the hand-crafted features for unsupervised learning. As the hand-crafted features have poor performance, various deep learning methods are proposed. \cite{Yu2017,Fan2017} assumed that the dataset has clustering properties. They belong to pseudo label learning. \cite{Deng2017,Zhong2017a} adopted CycleGAN \cite{Zhu2017} to align the distribution of source and target domains. Nevertheless, these methods usually rely on prior knowledge cues like camera information and are not easy to train. Our proposed TF-DIM model demands no extra knowledge and can be used as a supplement to other methods.

\subsection{Mutual-information Estimation}
MI has developed in the domain of feature learning for a long time. Linsker and Bell \cite{Linsker,Bell} applied the information principle to the neural network. Ji et al. \cite{Ji2018} maximized the MI between the class pairs for unsupervised clustering and segmentation. Contrastive Predictive Coding (CPC) \cite{Oord2018} learned useful representations from high-dimensional space by MI estimation. Mutual Information Neural Estimator (MINE) \cite{Belghazi2018} aimed at supervised classification problems by maximizing the MI. \cite{Hjelm2018} replaces the KL-based divergence with Jensen-Shannon divergence (JSD) as the latter is bound to better results. Inspired by \cite{Hjelm2018}, we explore some possibilities around ReID issue.

\begin{figure}
    \centering
    \includegraphics[width=2in]{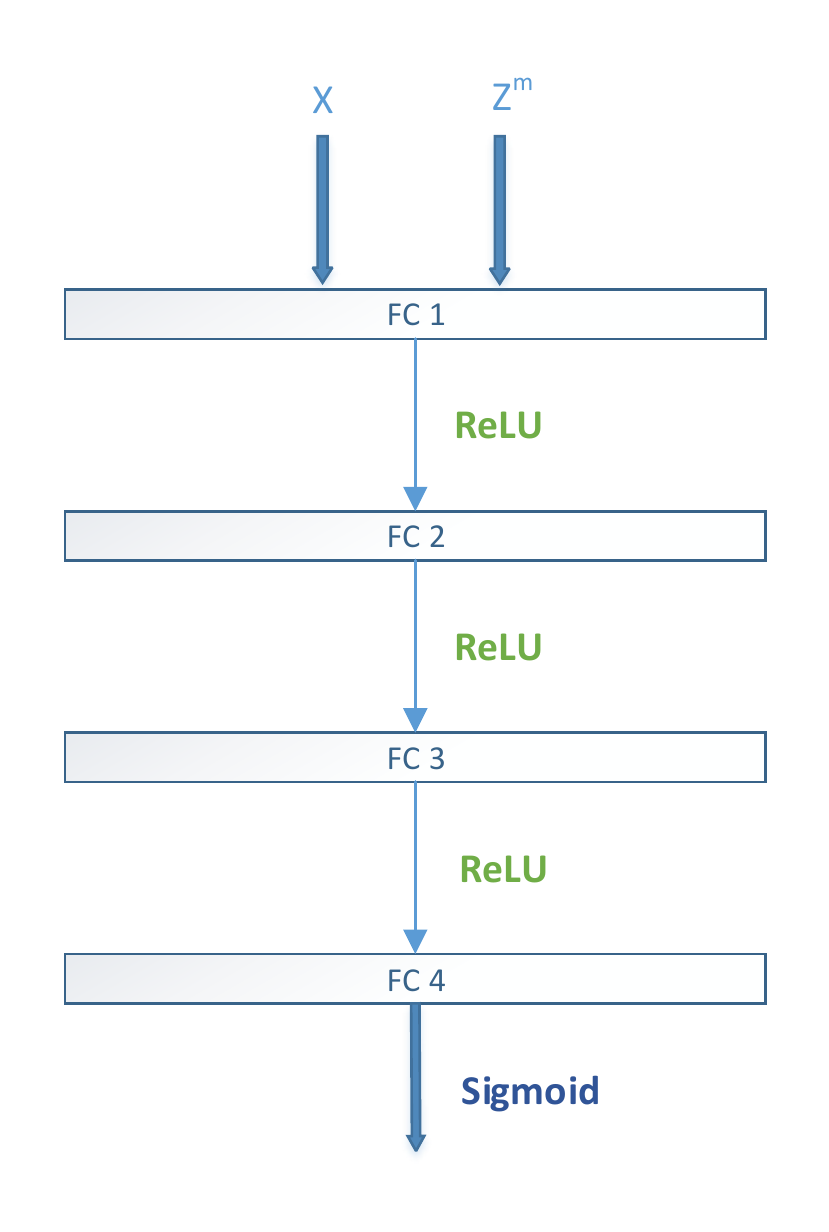}
    \caption{The structure of the DIM network.}
    \label{fig:my_label2}
\end{figure}

\section{Deep InfoMax Network}
The traditional works pay more attention to the encoder features and the ground-truth labels, in which the feature representations tend to preserve the commonality of categories. However, there are many commonalities between pedestrians in ReID task. In other words, the inter-class distance for pedestrians is not big enough and the intra-class distance is not small enough.

The discriminative feature representations in ReID task should have the ability to identify the pedestrian from the entire dataset, that is, extracting the unique information from the image. For this purpose, we maximize the MI between input pedestrians $X=\left\{x_{i}\right\}_{i=1}^{N}$ and encoder outputs $Z=\left\{z_{i}\right\}_{i=1}^{N}$, where $N$ is the number of images.

The MI between $X$ and $Z$ is defined as:

\begin{equation}
\label{eq1}
    \mathcal{L}_{DIM} = I(X,Z) = \iint p(z|x) \tilde{p}(x) log\frac{p(z|x)}{p(z)}dxdz
\end{equation}

where $ \tilde{p}(x)$ is the distribution of raw dataset $X$, and $p(z)$ is the distribution of $Z$.

\begin{equation}
    \label{eq2}
    p(z) = \int p(z|x) \tilde{p}(x)dx
\end{equation}

We transform the Eq.\ref{eq1} so as to compute the maximization of MI.
\begin{align}
    I(X,Z) &= \iint p(z|x) \tilde{p}(x) log \frac{p(z|x) \tilde{p}(x)}{p(z) \tilde{p}(x)}dxdz      \\
          &= D_{KL}(p(z|x) \tilde{p}(x)|| p(z) \tilde{p}(x))
    \label{eq3}
\end{align}
where $D_{KL}$ denotes the Kullback-Leibler divergence \cite{NIPS2016_6066}. The Eq.\ref{eq3} means that the maximization of MI is to enlarge the distance between $p(z|x)\tilde{p}(x)$ and $p(z)\tilde{p}(x)$. However, the upper limit $D_{KL}$ is unbounded, which is likely to achieve an infinite result. Therefore, we introduce a bounded Jensen-Shannon divergence $D_{JS}$ \cite{Fuglede2004} instead
\begin{equation}
    \label{eq4}
    D_{JS}(P,Q) = \frac{1}{2}D_{KL}(P|| \frac{P+Q}{2}) + \frac{1}{2}D_{KL}(Q||\frac{P+Q}{2})
\end{equation}
$P$ and $Q$ refer as $p(z|x)\tilde{p}(x)$ and $p(z)\tilde{p}(x)$ respectively.
Combined with Eq.\ref{eq4}, \ref{eq3} and \ref{eq1}, the object function is
\begin{equation}
    \label{eq5}
    \mathcal{L}_{DIM} =  \alpha \cdot D_{JS}(p(z|x) \tilde{p}(x), p(z)\tilde{p}(x))
\end{equation}
where $\alpha$ is a constant value.

In \cite{Nowozin2016}, $D_{JS}$ can be concluded as
\begin{equation}
    \label{eq6}
    \begin{split}
    D_{JS}(P,Q) = & \max_{D_{\omega}}  \mathbb E_{x \backsim p(z|x)\tilde{p}(x)}[log D_{\omega}(x)] \\
    & + \mathbb E_{x \backsim  p(z)\tilde{p}(x)}[log(1-D_{\omega}(x))]
    \end{split}
\end{equation}
where $D_{\omega}$ is the DIM network which will be discussed in section \ref{section3.3}. Finally, bring Eq.\ref{eq6} into \ref{eq5}, the object function is
\begin{equation}
\label{eq7}
\begin{split}
    \mathcal{L}_{DIM} = &\max_{D_\omega} \{ \alpha \cdot (\mathbb E_{x \backsim p(z|x)\tilde{p}(x)}[log D_{\omega}(x)] \\
    &+ \mathbb E_{x \backsim p(z)\tilde{p}(x)}[log(1-D_{\omega}(x))]) \}
\end{split}
\end{equation}


\label{section3.3}

In order to perform the Deep InfoMax loss in eq.\ref{eq10}, we construct the DIM network referred as the $D_{\omega}$. The structure of the DIM network is demonstrated in Fig.\ref{fig:my_label2}. There are four FC layers, first three FC layers followed by ReLU activation and last FC layer followed by sigmoid activation. 

The DIM network needs positive and negative samples as input. There are two available sampling strategies. The first idea is to choose $(x,z)$, $(x',z)$ as sample pairs, where $x'$ denotes images randomly disrupted in a batch \cite{Hjelm2018}. The second solution is partly sampling which introduces the ground-truth identity labels. Although the first strategy seems to give wrong negative samples for the DIM network, the experiments in section \ref{section5.2} show that random sampling has competitive performance over prior labels strategy. 

In this paper, we choose a random sampling strategy for the DIM network. According to Eq.\ref{eq7}, The DIM network loss is rewritten as
\begin{equation}
    \label{eq8}
    \begin{split}
    \mathcal{L}_{DIM}((x,z),(x',z))=& - \min_{D_\omega} \{ \alpha \cdot \mathbb E_{(x,z)\backsim p(z|x)\tilde{p}(x)}[log D_{\omega}(x,z)] \\
    &+ \mathbb E_{(x',z)\backsim p(z)\tilde{p}(x)}[log(1-D_{\omega}(x',z))] \}
    \end{split}
\end{equation}

\begin{figure*}
    \centering
    \includegraphics[width=5in]{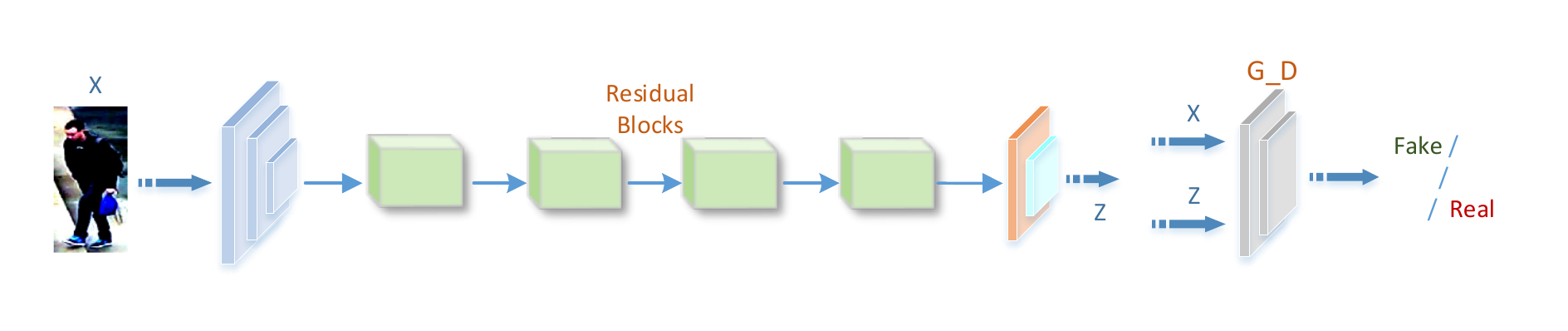}
    \caption{The framework of our Global-DIM network. $X$ and $Z$ represent the input image and the output of the encoder network respectively. $D_w$ denotes the DIM network.}
    \label{fig:my_label3}
\end{figure*}

\section{Global DIM Network}
Our approach aims to introduce Deep InfoMax loss to enhance the capability of feature representations in ReID task. Based on the Deep InfoMax network in the above, we propose our Global DIM network. The whole framework is shown in Fig.\ref{fig:my_label3}, which consists of two parts: encoder and DIM networks.

The encoder network is improved on the IDE baseline. The baseline uses ResNet-50 \cite{He2016} as backbone, which finetunes on the pre-trained model of ImageNet challenge \cite{Socher2009}. Following the strategy in \cite{Zhong}, we discard the original final 1000 dim fully connected layer (FC), add two FC layers: the output of the first layer is 512 dim, followed with Batch Normalization \cite{Ioffe2015}, LeakyReLU, and Dropout layer \cite{Srivastava2014}. The output of the second FC layer is ID dim, where ID is the number of identities in the datasets. 

The encoder network based on IDE baseline is an effective supervised ReID model. The cross-entropy loss employed in the encoder network is
\begin{equation}
    \label{eq9}
        \mathcal{L}_{cls} = -\frac{1}{N}\sum_{i=1}^{N}y_{i}log p(z_i) 
\end{equation}
where $N$ is the number of images, $y_i$ is the $i$-th label in a batch.


Based on the feature representations obtained from the encoder network, our Global DIM network considers the correlation between the input and the feature representations, in which adds the DIM network. The full objective loss is:

\begin{equation}
    \label{eq10}
    \mathcal{L}_{Global-DIM}= \mathcal{L}_{cls}+\upbeta \mathcal{L}_{DIM}((x,z),(x',z))
\end{equation}
where $\upbeta$ is hyper-parameter which denote the weight of the DIM loss in the whole objective loss.

\section{Local DIM Network}
\begin{figure*}
    \centering
    \includegraphics[width=5in]{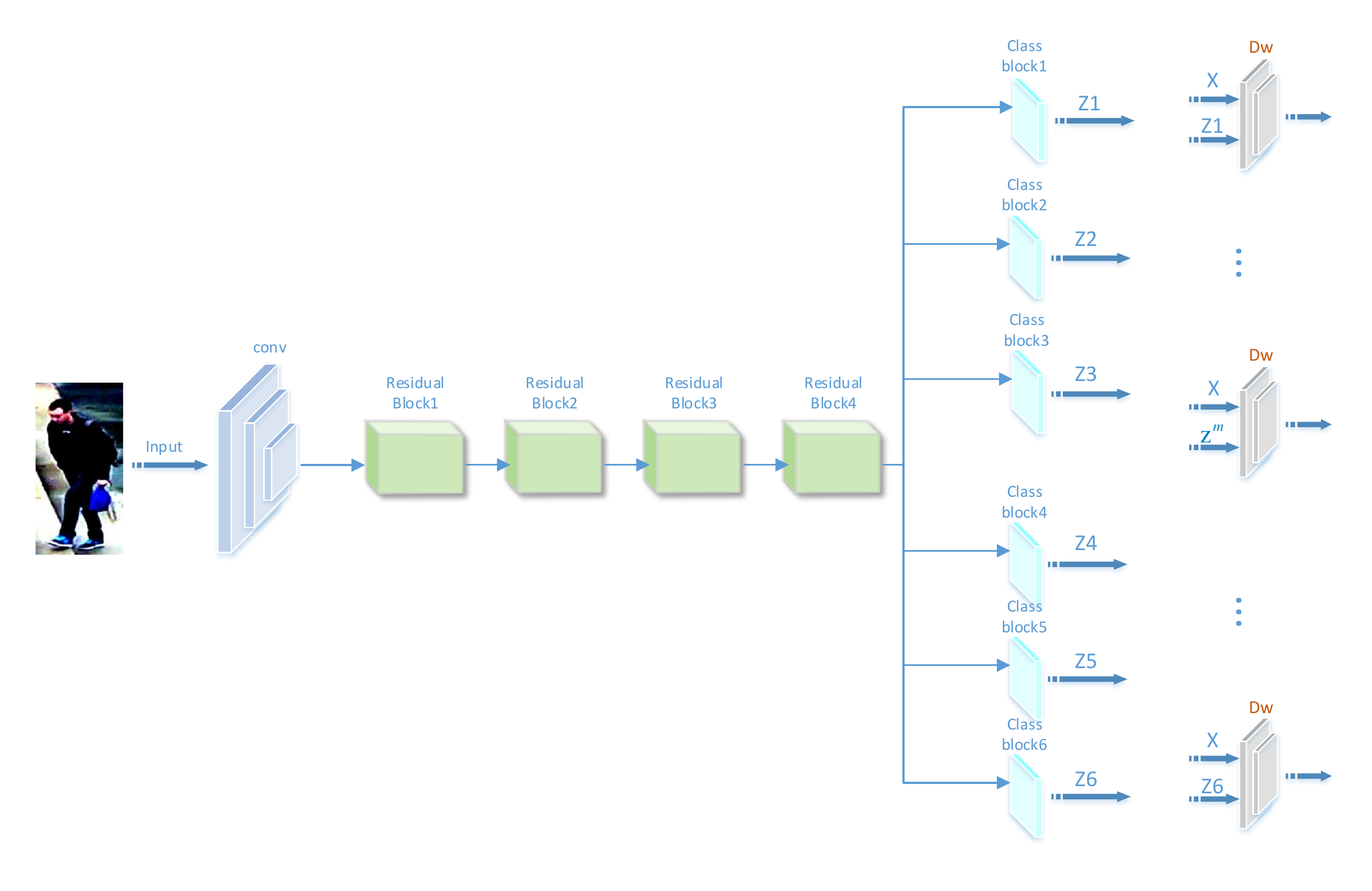}
    \caption{The framework of our Local-DIM network. $X$ and $Z$ represent the input image and the output of the encoder network respectively. $D_w$ denotes the DIM network.}
    \label{fig:my_label4}
\end{figure*}

Inspired by \cite{Hjelm2018} that the DIM method takes effects in local features, we propose our Local DIM approach based on the local features. The framework of the Local-DIM network is demonstrated in Fig. \ref{fig:my_label4}, consisting of two parts: part encoder and DIM networks.

Our part encoder network is improved on the PCB model. The encoder network uses ResNet-50 \cite{He2016} as the backbone, which is finetuned on the pre-trained model of ImageNet challenge \cite{Socher2009}. After the global average pooling, the 3D tensor block is partitioned into six branches, which represent the six parts of the pedestrian. Then, each branch is followed with a classifier block: the output of the first FC layer is 512 dim, trailed by Batch Normalization \cite{Ioffe2015}, LeakyReLU, and Dropout layer \cite{Srivastava2014} in sequence. The output dimension of the second FC layer equals the number of identities in the dataset. 

The part encoder model is an effective method to identify the pedestrian given the datasets in supervised learning. In the part encoder model, the pedestrian is partitioned into six parts for extracting important part features. Then, the cross-entropy loss is employed for important part features to ensure the effective feature representations.

The loss of the part encoder network $\mathcal{L}_{P}$ is
\begin{equation}
    \label{eq11}
    \mathcal{L}_{P} = -\frac{1}{N}\sum_{m=1}^{M}\sum_{i=1}^{N}y_{i}log p(z_i^m)
\end{equation}
where $M$ is the number of branches, $z_i^m$ denotes the $m$-th part feature of the $i$-th image, and $p(z_i^m)$ is the probability distribution of $z_i^m$, $y_i$ is the label corresponding to $x_i$.

In summary, we propose an end-to-end Local-DIM model based on the part encoder and DIM networks. The full objective loss is:
\begin{equation}
    \label{eq12}
    \mathcal{L}_{Local-DIM}= \mathcal{L}_{P}+\lambda \sum_{m=1}^{M} \mathcal{L}_{DIM}((x,z^m),(x',z^m))
\end{equation}
where $\lambda$ is hyper-parameter which denote the weight of the DIM loss in the whole objective loss. 

\section{Unsupervised Transfer Learning with DIM}

Our approach provides a solution to address the cross-dataset unsupervised ReID issue as a result of the fact that the DIM network needs no extra labels information. The cross-dataset unsupervised method is called TF-DIM. The TF-DIM transfers the feature representations from a labeled source dataset and finetunes on the unlabeled target dataset. The concrete steps are:

$\bullet$ \textbf{step(1): Supervised part-based DIM learning in the labeled source dataset.} As discussed above, we train the Global and Local-DIM networks for learning an encoder network which has discriminative feature representations.

$\bullet$ \textbf{step(2): Transfer model in the unlabeled target dataset.} In this step, we load the weight of the encoder network to the unlabeled target dataset. The DIM network is initialized to adapt the unlabeled target dataset.

$\bullet$ \textbf{step(3): Finetune on the unlabeled target dataset.} Based on the pretrained model, we finetune the unlabeled dataset with the DIM loss.

\section{Implementation}

\textbf{Training for Global-DIM network} The Global-DIM network involves the global encoder and DIM networks. The global encoder generates the global feature representations, and the DIM network maximizes the MI between the feature representations and input pedestrians. Because the input pedestrians and feature representations mismatch in size, the input pedestrians $x$ and $x'$ are reduced to $z$ and $z'$ in practice through the global encoder network. $z'$ is the output of the global encoder network corresponding to the input $x'$. The Global-DIM network is optimised with SGD optimizer in 60 epochs. The learning rate is 0.3, decaying by 10 after 40 epochs. The input images are resized to 256x128. 

\textbf{Training for Local-DIM network} The Local-DIM network consists two parts: the local encoder and six DIM networks. The encoder network takes input images and outputs the six important part features. The six part features correspond to six DIM networks, and the sampling strategy in each DIM network is the same as the Global-DIM method. The weights of the six DIM networks are shared for learning the correlations between six parts. The input pedestrians are resized to 384x192. The whole Local-DIM network is trained with SGD optimizer in 60 epochs. The learning rate sets to 0.02, decaying by 10 after 40 epochs.
 
\textbf{Training for TF-DIM method} The TF-DIM method is built on the basis of Local-DIM framework. The Local-DIM model trained on the source dataset is served as a pre-trained model for transfer learning in the target dataset. The DIM network dominates the cross-dataset TF-DIM approach because of its unsupervised DIM loss. We finetune on the target dataset with SGD optimizer in 60 epochs. The learning rate is 0.00005, decaying by 10 after 40 epochs.

\section{Experiments}

\subsection{Datasets}

Our approach is evaluated on three ReID datasets: Market-1501 \cite{Zheng}, DukeMTMC-reID \cite{Zheng2017,Ristani2016} and CUHK03 \cite{Li}.

\noindent
\textbf{Market-1501} contains 32668 annotated images of 1501 identities from six overlapped cameras. There are 12936 images of 751 identities for training, 19732 images of 750 identities for testing. The query set collects 3368 images of 750 identities to search in the gallery set.

\noindent
\textbf{DukeMTMC-reID} consists of 36411 bounding boxes of 1404 identities. During the evaluation phase, there are 16522 images of 702 identities for training, 17661 images of the other 702 identities for testing. The 2208 query images are picked from the gallery set.

\noindent
\textbf{CUHK03} has 14096 images of 1476 identities from two cameras. The dataset consists of labeled bounding boxes and DPM detected bounding boxes. In our experiments, we use the detected bounding boxes as it is closer to practical application. We follow the new protocol \cite{Zhong2017} to evaluate our approach. 

\subsection{Evaluation}
\label{section5.2}

\begin{figure}[!t]
    \centering
    \includegraphics[width=3in]{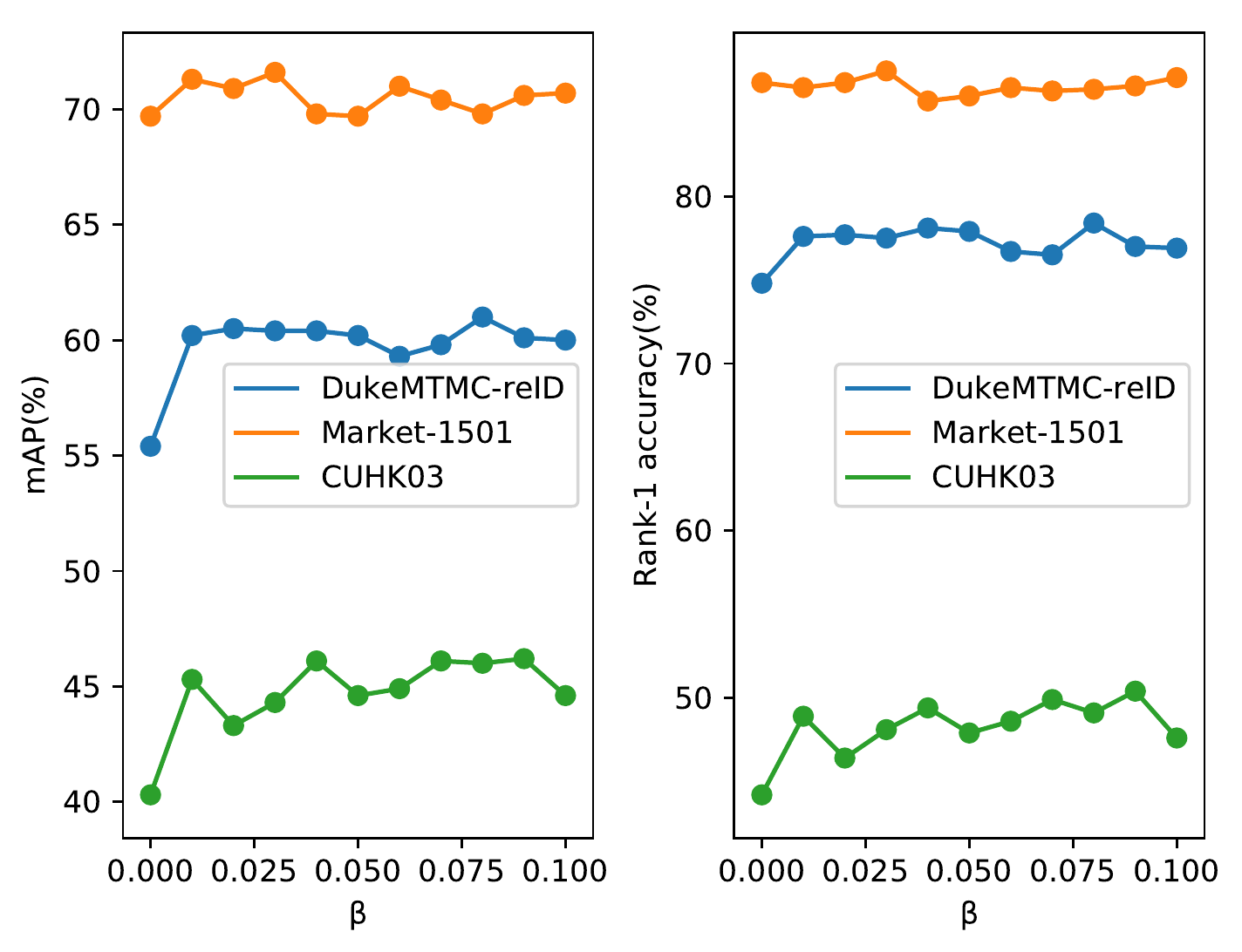}
    \caption{The evaluation about hyper-parameter $\beta$ in Eq. \ref{eq10}.}
    \label{fig:my_label5}
\end{figure}

\begin{figure}[!t]
    \centering
    \includegraphics[width=3in]{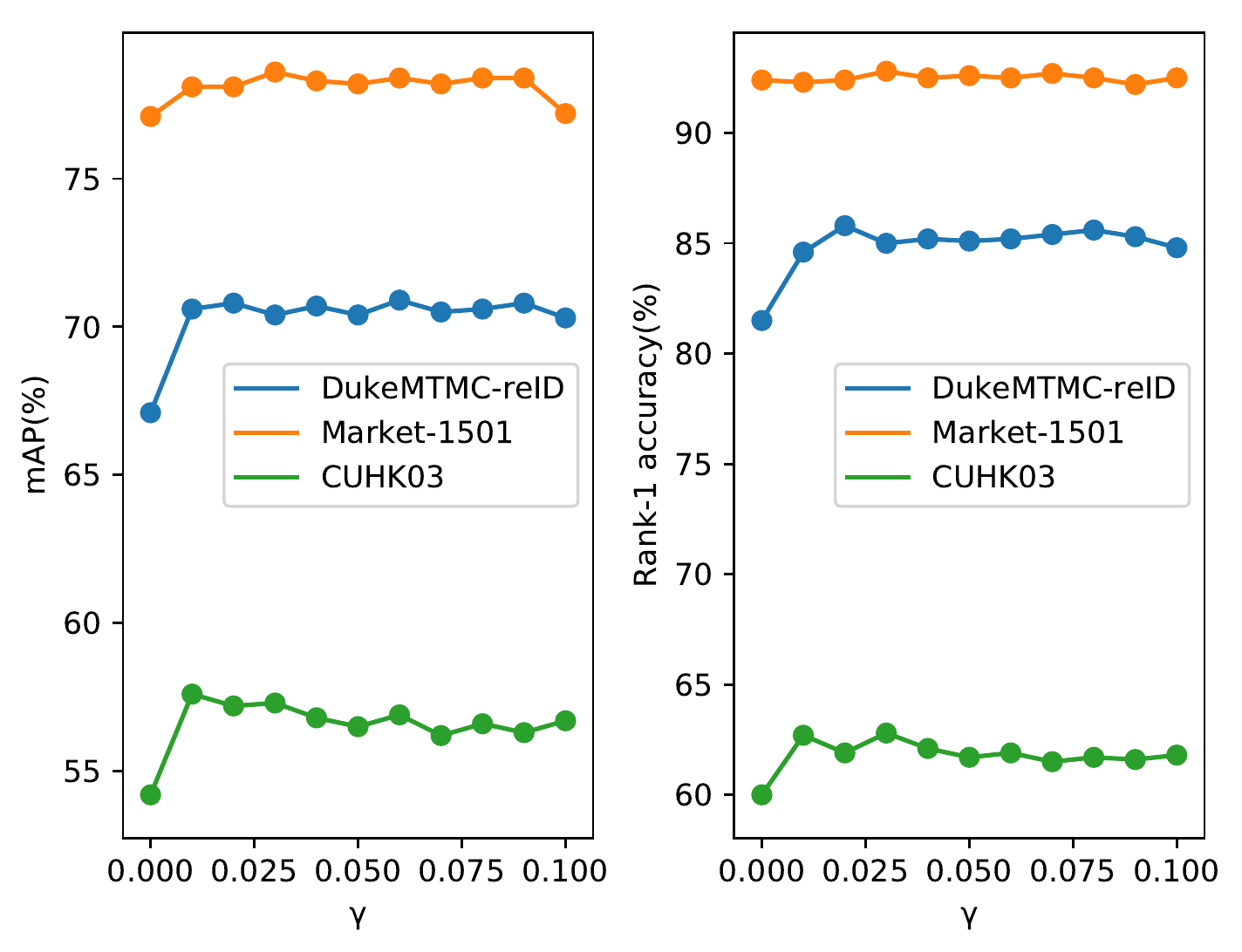}
    \caption{The evaluation about hyper-parameter $\lambda$ in Eq. \ref{eq12}.}
    \label{fig:my_label6}
\end{figure}

\begin{figure}[!t]
    \centering
    \includegraphics[width=3in]{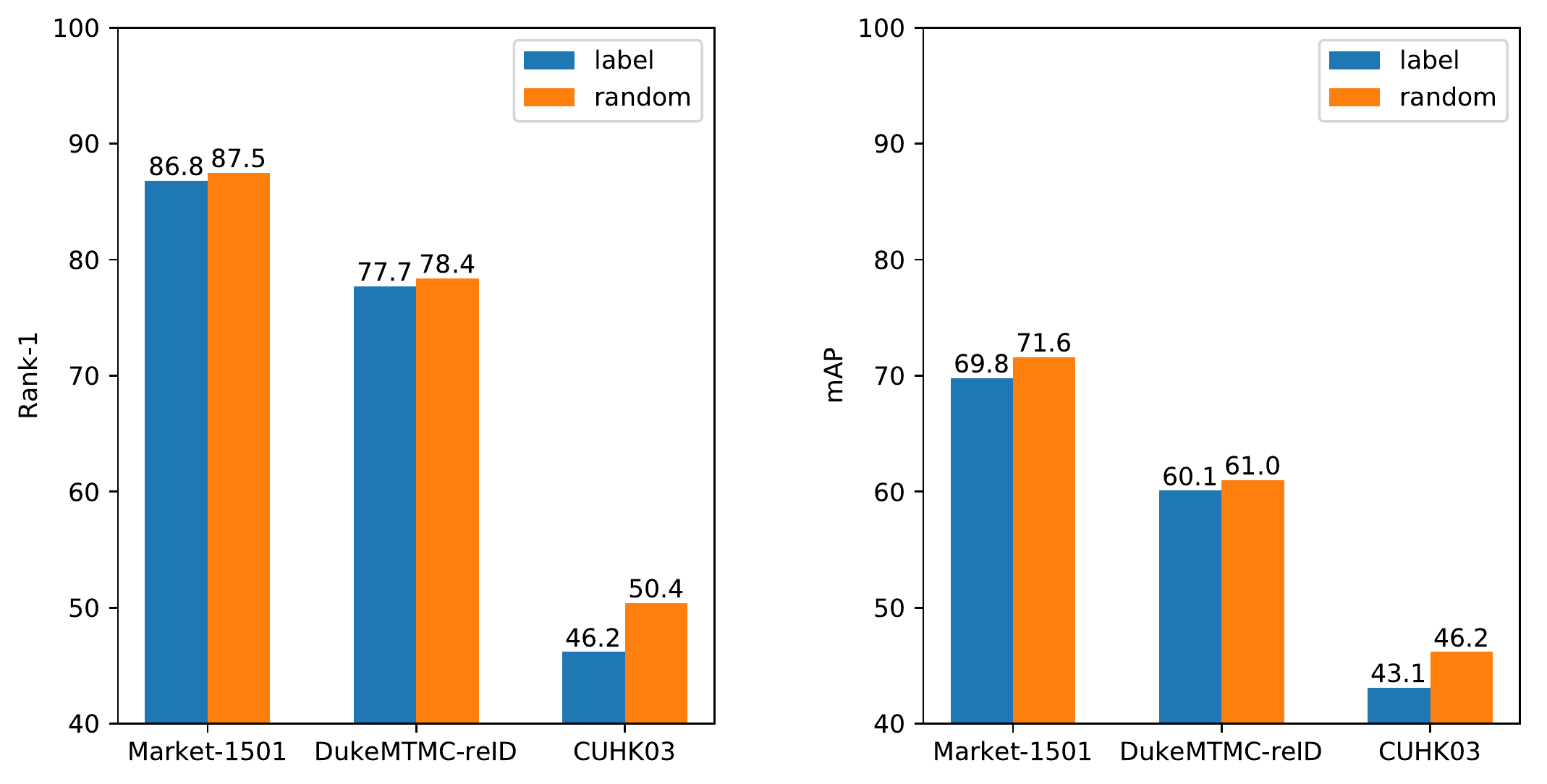}
    \caption{Comparison between the random and label sampling strategies in Global-DIM network.}
    \label{fig:my_label7}
\end{figure}

\begin{figure}[!t]
    \centering
    \includegraphics[width=3in]{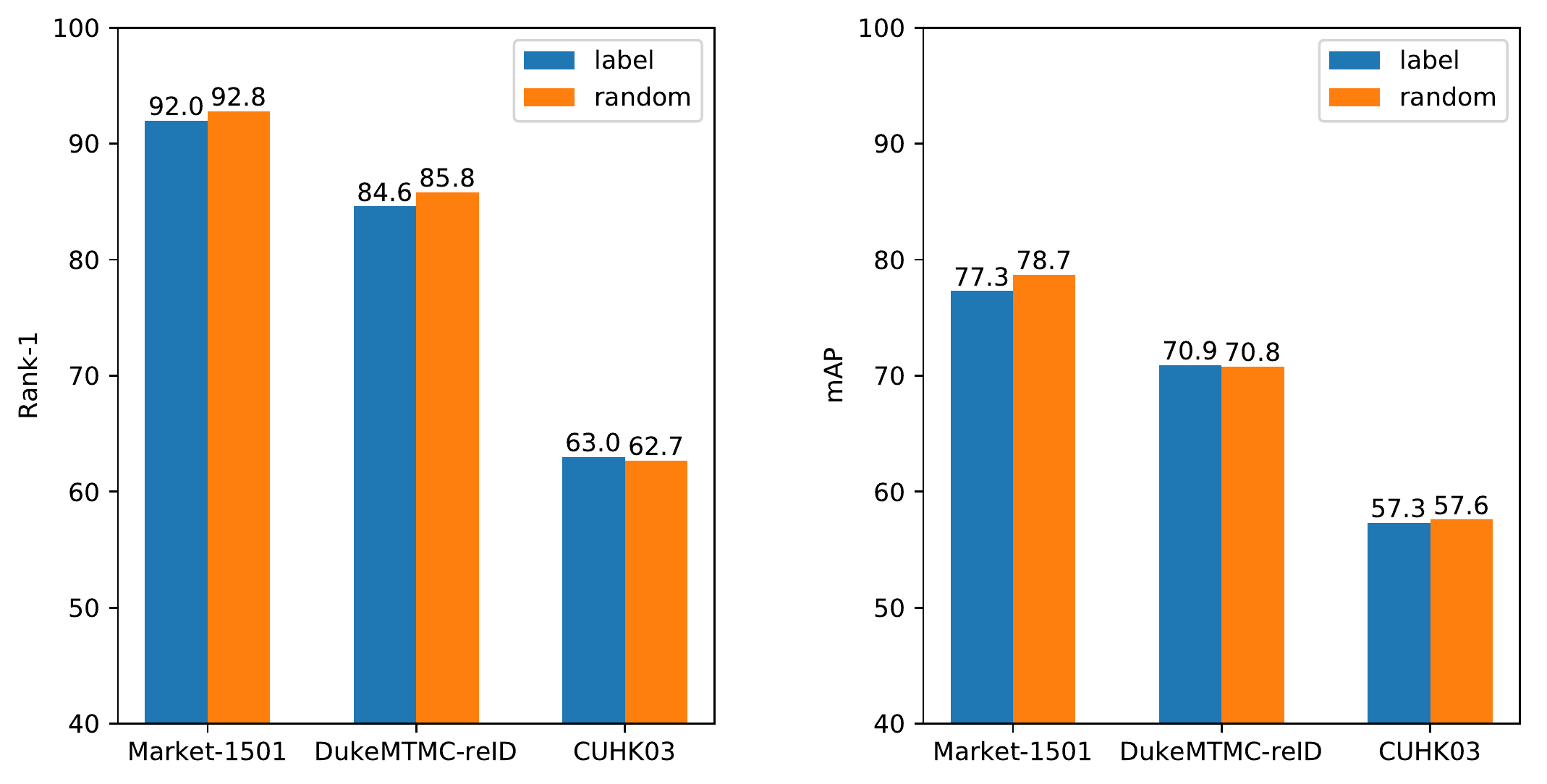}
    \caption{Comparison between the random and label sampling strategies in Local-DIM network.}
    \label{fig:my_label8}
\end{figure}

\textbf{Important Parameters.} We evaluate the hyper-parameters $\beta$ and $\lambda$ to demonstrate the effectiveness of DIM loss. The results are shown in Fig.\ref{fig:my_label5} and \ref{fig:my_label6}. Our approach improves obviously over the IDE and PCB baseline respectively, which proves the validity of DIM loss. In the Global-DIM network, the mAP and Rank-1 achieve the best performance when $\beta$ equals to 0.02. At the same time, the optimal $\lambda$ is 0.01 for the Local-DIM network to achieve better performance.

\textbf{Comparison between the random and label sampling.} The label sampling introduces the ground-truth identity labels for the DIM network to form input pairs. The performance of the random and label sampling strategies is demonstrated in Fig.\ref{fig:my_label7} and \ref{fig:my_label8}. To some extent, random sampling is superior to the label one. It is because that when sampling a few images from the dataset, the possibility of choosing the images belonging to the same identity is relatively low. Also, the random sampling strategy has an advantage in unsupervised learning. Therefore, we choose the random sampling strategy for the DIM network.

\begin{figure}[!t]
    \centering
    \includegraphics[width=3in]{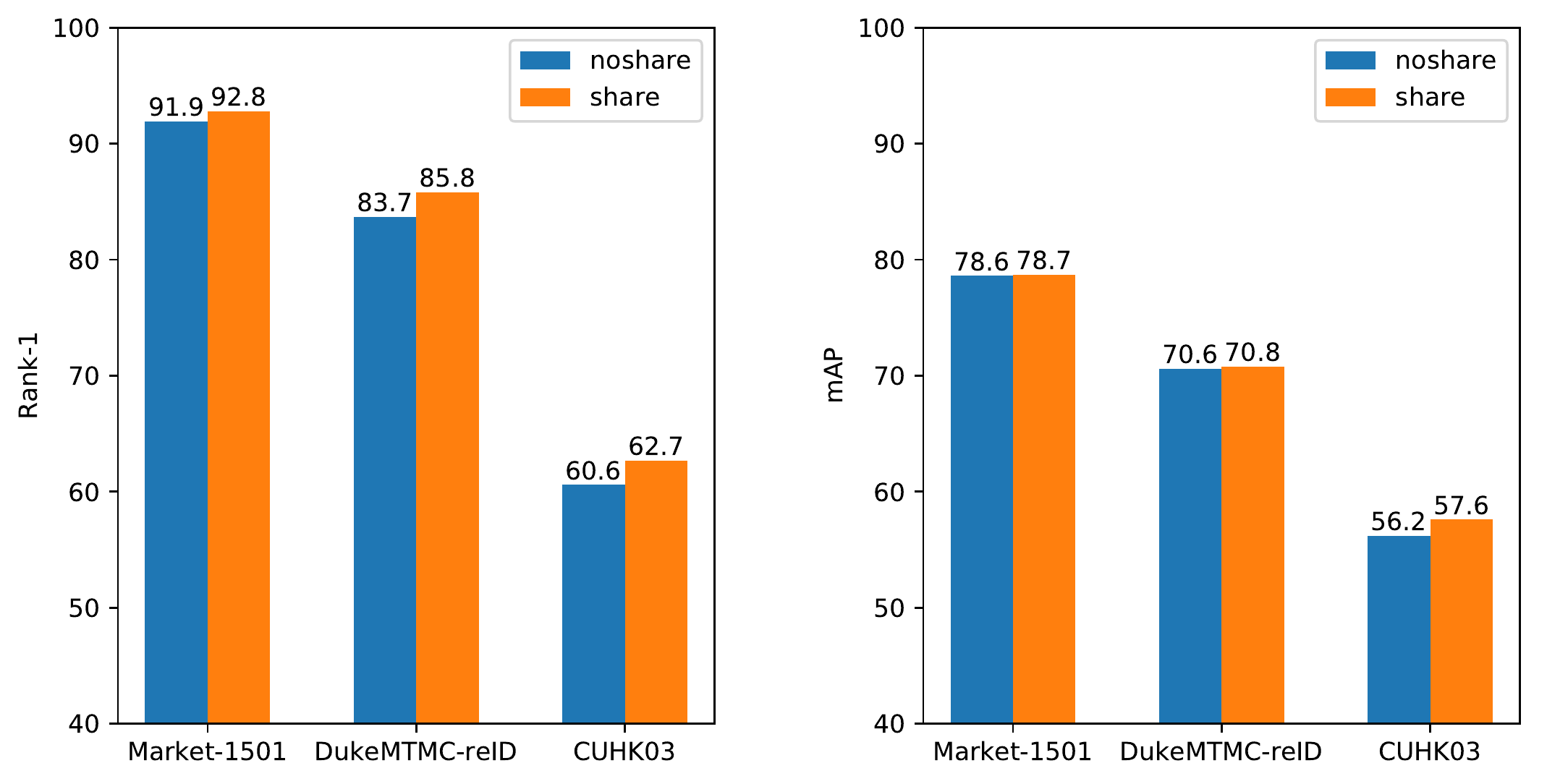}
    \caption{Comparison between the parameters sharing of six DIM networks in the Local-DIM method. The noshare in the chart refers that the parameters in the six DIM network are not shared. And the share means shared parameters.}
    \label{fig:my_label9}
\end{figure}

\textbf{Comparison between the parameters sharing of six DIM networks in the Local-DIM method.} There are six DIM networks in our Local-DIM model. Whether the six DIM networks share parameters is nontrivial because of the time-consuming problem in the training stage. In Fig. \ref{fig:my_label9}, we compare the results of shared and not shared parameters in six DIM networks. The empirical results show that the parameters sharing strategy is superior to the non-sharing ones. The underlying theory may be that the six DIM networks are correlated by parameters sharing strategy, achieving discriminative part features.

\textbf{Comparison between the IDE baseline and Global-DIM model.} The experiments between the IDE and Global-DIM model are specified in Table \ref{tab:1}. The Global-DIM model gains +0.7\% and +1.9\% improvements in rank-1 and mAP accuracy on Market-1501 respectively. Also, the Global-DIM has +3.6\% and +5.6\% improvements on DukeMTMC-reID. When tested on CUHK03, the Global-DIM increases +6.2\% and +5.9\% in rank-1 and mAP. The above experiments prove the validity of the Global-DIM model. Also, compared with the performance in DukeMTMC-reID and CUHK03 datasets, the improvements in Market-1501 dataset is a little weak. This may be because there are fewer pedestrian variances in the Market-1501 than DukeMTMC-reID and CUHK03, which is consistent with our motivation.

\begin{table}[]
    \centering
    \caption{Comparison of the IDE baseline and our Global-DIM models on three datasets: Market-1501, DukeMTMC-reID and CUHK03.}
    \begin{tabular}{c|c|c|c|c|c}
    \hline
        Datasets & Methods & rank-1 & rank5 & rank10 & mAP\\
    \hline
    \multirow{2}{*}{Market}
         & IDE & 86.8 & \textbf{95.2} & \textbf{96.9} & 69.7 \\ \cline{2-6}
    
         & Global-DIM & \textbf{87.5} & 95.0 & 96.7 & \textbf{71.6} \\ \cline{2-6}
         
    \hline
    \multirow{2}{*}{Duke}
         & IDE & 74.8 & 86.4 & 90.2 & 55.4 \\ \cline{2-6}
        
         & Global-DIM & \textbf{78.4} & \textbf{88.9} & \textbf{91.2} & \textbf{61.0} \\ \cline{2-6}
        
    \hline
    \multirow{3}{*}{cuhk03}
         & IDE & 44.2 & 65.4 & 73.5 & 40.3 \\ \cline{2-6}
        
         & Global-DIM & \textbf{50.4} & \textbf{68.9} & \textbf{77.9} & \textbf{46.2} \\ \cline{2-6}

    \hline
    \end{tabular}
    
    \label{tab:1}
\end{table}

\begin{table}[]
    \centering
    \caption{Comparison of the PCB baseline and our Local-DIM model on three datasets: Market-1501, DukeMTMC-reID and CUHK03. The performance in PCB model is evaluated by ourself, which is little different from the benchmark.}
    \begin{tabular}{c|c|c|c|c|c}
    \hline
        Datasets & Methods & rank-1 & rank5 & rank10 & mAP\\
    \hline
    \multirow{2}{*}{Market}
         & PCB & 92.4 & 97.2 & \textbf{98.2} & 77.1 \\ \cline{2-6}
    
         & Local-DIM & \textbf{92.8} & \textbf{97.3} & 98.1 & \textbf{78.7} \\ \cline{2-6}
    \hline
    \multirow{2}{*}{Duke}
         & PCB & 81.5 & 89.3 & 91.6 & 67.1 \\ \cline{2-6}
        
         & Local-DIM & \textbf{85.8} & \textbf{92.9} & \textbf{94.5} & \textbf{70.8} \\ \cline{2-6}
    \hline
    \multirow{2}{*}{cuhk03}
         & PCB & 60.0 & 78.3 & 84.9 & 54.2 \\ \cline{2-6}
        
         & Local-DIM & \textbf{62.7} & \textbf{79.9} & \textbf{86.2} & \textbf{57.6} \\ \cline{2-6}
    \hline
    \end{tabular}
    
    \label{tab:2}
\end{table}

\textbf{Comparison between the PCB baseline and Local-DIM model.} The performance between the PCB and Local-DIM model is demonstrated in Table \ref{tab:2}. On Market-1501, the Local-DIM model gains +0.4\% and +1.6\% in rank-1 accuracy and mAP. Also, the performance increases +4.3\% and +3.7\% respectively on DukeMTMC-reID. When evaluated on CUHK03 dataset, the results get +2.7\% and +3.4\% increase. The experiments in Table \ref{tab:1} and \ref{tab:2} prove the generality of our DIM network in global and local feature representations.

\textbf{Comparison between the Direct Transfer and TF-DIM model based on Global-DIM network.} The Direct Transfer method means that the target dataset is directly tested on the model trained on the source dataset. The encoder network is based on the Global-DIM network trained on the source dataset. Table \ref{tab:3} illustrates the comparison results between the Direct Transfer and the TF-DIM model. When the DukeMTMC-reID is the source dataset and Market-1501 is the target dataset, our TF-DIM method has +0.8\% and +0.6\%  gains in rank-1 accuracy and mAP. When we test the model on the DukeMTMC-reID dataset, our approach gains +2.7\% and +2.2\% in rank-1 accuracy and mAP.

\textbf{Comparison between the Direct Transfer and TF-DIM model based on Local-DIM network.} The Table \ref{tab:4} shows the comparison between the TF-DIM model and the Direct Transfer method. The encoder network is based on the Local-DIM network trained on the source dataset. When the DukeMTMC-reID is the source dataset and Market-1501 is the target dataset, our TF-DIM method gains +1.7\% and +1.2\% in rank-1 accuracy and mAP. The improvements are +2.6\% and +1.7\% in rank-1 accuracy and mAP when the Market-1501 is the source dataset.

\begin{table}[]
    \centering
    \caption{Comparison of our TF-DIM model with the Direct Transfer method on DukeMTMC-reID and CUHK03 datasets. The encoder network is based on the Global-DIM network.}
    \label{tab:3}
    \begin{tabular}{c|c|c|c|c}
    \hline
    \multirow{2}{*}{Methods}
           &  \multicolumn{2}{c|}{Duke->Market1501} & \multicolumn{2}{c}{Market-1501->Duke} \\ \cline{2-5}
           & rank-1 & mAP & rank-1 & mAP \\ \cline{2-5}
    \hline
    Direct Transfer & 46.4 & 20.1 & 32.8 & 17.3 \\
    TF-DIM & \textbf{47.2} & \textbf{20.7} & \textbf{35.5} & \textbf{19.5} \\
    \hline
    \end{tabular}
    
\end{table}

\begin{table}[]
    \centering
    \caption{Comparison of our TF-DIM model with the Direct Transfer method on DukeMTMC-reID and CUHK03 datasets. The encoder network is based on the Local-DIM network.}
    \label{tab:4}
    \begin{tabular}{c|c|c|c|c}
    \hline
    \multirow{2}{*}{Methods}
           &  \multicolumn{2}{c|}{Duke->Market1501} & \multicolumn{2}{c}{Market-1501->Duke} \\ \cline{2-5}
           & rank-1 & mAP & rank-1 & mAP \\ \cline{2-5}
    \hline
    Direct Transfer & 56.4 & 27.2 & 40.1 & 22.8 \\
    TF-DIM & \textbf{58.0} & \textbf{28.4} & \textbf{42.7} & \textbf{24.5} \\
    \hline
    \end{tabular}
    
\end{table}

\begin{table}[]
    \centering
    \caption{Comparison of our Global-DIM and Local-DIM networks with the state-of-art methods on Market-1501. The compared methods are divided into three categories: hand-crafted ReID, deep global feature learning and deep part feature learning methods.}
    \label{tab:5}
    \begin{tabular}{c|c|c|c|c}
    \hline
          Methods & rank-1 & rank5 & rank10 & mAP \\
    \hline
          BoW+Kissme \cite{Zheng} & 44.4 & 63.9 & 72.2 & 20.8 \\ 

          WARCA \cite{Jose2016} & 45.2 & 68.1 & 76.0 & - \\ 
    
          KLFDA \cite{Karanam2016} & 46.5 & 71.1 & 79.9 & - \\
    \hline
          SOMAnet \cite{Barbosa2018} & 73.9 & - & - & 47.9 \\ 
        
          SVDNet \cite{Sun2017a} & 82.3 & 92.3 & 95.2 & 62.1 \\ 
        
          PAN \cite{Zheng2018} & 82.8 & - & - & 63.4 \\
          
          Transfer \cite{Geng2016} & 83.7 & - & - & 65.5 \\
    \hline
          MultiRegion \cite{Ustinova2015} & 66.4 & 85.0 & 90.2 & 41.2 \\ 
        
          HydraPlus \cite{Liu2017} & 76.9 & 91.3 & 94.5 & - \\ 

          PAR \cite{Zhao2017} & 81.0 & 92.0 & 94.7 & 63.4 \\
          
          MultiLoss \cite{Li2017} & 83.9 & - & - & 64.4 \\
          
          PDC* \cite{Su2017} & 84.4 & 92.7 & 94.9 & 63.4 \\
          
          PartLoss \cite{Yao2017} & 88.2 & - & - & 69.3 \\
          
          MultiScale \cite{Chen} & 88.9 & - & - & 73.1 \\
    \hline
          IDE & 86.8 & 95.2 & 96.9 & 69.7 \\
          Global-DIM & \textbf{87.5} & 95.0 & 96.7 & \textbf{71.6} \\
    \hline
          PCB & 92.4 & 97.2 & \bf{98.2} & 77.1 \\
          Local-DIM & \bf{92.8} & \bf{97.3} & 98.1 & \bf{78.7} \\
    \hline
    \end{tabular}
    
\end{table}

\subsection{Comparison with the state-of-the-art methods}

\begin{table}[]
    \centering
    \caption{Comparison of our Global-DIM and Local-DIM models with the state-of-art methods on DukeMTMC-reID and CUHK03 datasets.}
    \label{tab:6}
    \begin{tabular}{c|c|c|c|c}
    \hline
    \multirow{2}{*}{Methods}
           &  \multicolumn{2}{c|}{DukeMTMC-reID} & \multicolumn{2}{c}{CUHK03} \\ \cline{2-5}
           & rank-1 & mAP & rank-1 & mAP \\ \cline{2-5}
    \hline
    BoW+Kissme \cite{Zheng} & 25.1 & 12.2 & 6.4 & 6.4 \\
    LOMO+XQDA \cite{Liao} & 30.8 & 17.0 & 12.8 & 11.5 \\
    GAN \cite{Zheng2017} & 67.7 & 47.1 & - & - \\
    PAN \cite{Zheng2018} & 71.6 & 51.5 & 36.3 & 34.0 \\
    SVDNet \cite{Sun2017a} & 76.7 & 56.8 & 41.5 & 37.3 \\
    MultiScale \cite{Chen} &79.2 & 60.6 & 40.7 & 37.0 \\
    TriNet+Era \cite{Zhong2017b} & 73.0 & 56.6 & 55.5 & 50.7 \\
    \hline
    IDE & 74.8 & 86.4 & 90.2 & 55.4 \\
    Global-DIM & \textbf{78.4} & \textbf{88.9} & \textbf{91.2} & \textbf{61.0} \\
    \hline
    PCB & 81.5 & 67.1 & 60.0 & 54.2 \\
    Local-DIM & \bf{85.8} & \bf{70.8} & \bf{62.7} & \bf{57.6} \\
    \hline
    \end{tabular}
    
\end{table}

\textbf{Comparison between our approach and the state-of-the-art models.}
We compare our approach with the state-of-the-art ReID methods. The results are shown in Table \ref{tab:5},\ref{tab:6}. Compared with the IDE and PCB baseline, our approach has considerable improvements in rank-1 accuracy and mAP. On Market-1501 dataset, the Local-DIM model gains +0.4\% and +1.6\% in rank-1 accuracy and mAP. Also, the performance increases +4.3\% and +3.7\% respectively on DukeMTMC-reID. When evaluated on CUHK03 dataset, the results get +2.7\% and +3.4\% increase. All the experiments are evaluated under single-query and no re-ranking strategy. In our paper, we achieve the state-of-the-art results on these three datasets.

\textbf{Comparison between the TF-DIM model and the state-of-the-art unsupervised domain adaptation methods.}
We compare our TF-DIM model with the state-of-the-art unsupervised ReID methods. The results are shown in Table \ref{tab:7}. DukeMTMC-reID/Market-1501 is the source dataset and Market-1501/DukeMTMC-reID is the target dataset. The compared methods are divided into three categories: hand-crafted, unsupervised and unsupervised domain adaptation methods. The Direct Transfer is directly tested on the target dataset based on our Global-DIM and Local-DIM methods. The results show that our TF-DIM model achieves competitive performance compared with related state-of-the-art methods.

\subsection{Discussion}
Although our TF-DIM method in Table \ref{tab:7} has a promising performance, it does not seem to be convincing in rank-1 accuracy and mAP. The key of the unsupervised domain adaptation ReID issue is to converge the distribution of the source and target datasets like \cite{Deng2017},\cite{Wei2017}. Our TF-DIM method proposes a DIM loss to enhance the feature representations, which can be a complement to unsupervised ReID methods. We want to emphasize that although our TF-DIM model has good universality, unsupervised ReID task needs to be further optimized in conjunction with datasets.

\begin{table}[]
    \centering
    \caption{Comparison of our TF-DIM model with the state-of-art methods on DukeMTMC-reID and CUHK03 datasets.}
    \label{tab:7}
    \begin{tabular}{c|c|c|c|c}
    \hline
    \multirow{2}{*}{Methods}
           &  \multicolumn{2}{c|}{Duke->Market1501} & \multicolumn{2}{c}{Market-1501->Duke} \\ \cline{2-5}
           & rank-1 & mAP & rank-1 & mAP \\ \cline{2-5}
    \hline
    LOMO \cite{Zheng} & 27.2 & 8.0 & 12.3 & 4.8 \\
    BoW \cite{Liao} & 35.8 & 14.8 & 17.1 & 8.3 \\
    \hline
    UMDL \cite{Peng}  & 34.5 & 12.4 & 18.5 & 7.3 \\
    PUL \cite{Fan2017} & 45.5 & 20.5 & 30.0 & 16.4 \\
    CAMEL \cite{Yu2017} & 54.5 & 26.3 & - & - \\
    \hline
    PTGAN \cite{Wei2017} & 38.6 & - & 27.4 & - \\
    SPGAN \cite{Deng2017} & 51.5 & 22.8 & 41.1 & 22.3 \\
    SPGAN+LMP \cite{Deng2017} & 57.7 & 26.7 & \bf{46.4} & \bf{26.2} \\
    TJ-AIDL \cite{Wang2018} & \bf{58.2} & 26.5 & 44.3 & 23.0 \\
    \hline
    Direct Transfer (Global-DIM) & 46.4 & 20.1 & 32.8 & 17.3 \\
    TF-DIM (Global-DIM) & 47.2 & 20.7 & 35.5 & 19.5 \\
    \hline
    Direct Transfer (Local-DIM) & 56.4 & 27.2 & 40.1 & 22.8 \\
    TF-DIM (Local-DIM) & 58.0 & \bf{28.4} & 42.7 & 24.5 \\
    \hline
    \end{tabular}
    
\end{table}

\section{Conclusion}

In this paper, we propose a DIM network which maximizes the MI between the input image and encoder output to boost the capability of feature representations in ReID issue. Taking advantage of the MI theory, the DIM network enhances the representative features from input pedestrians. The Global-DIM and Local-DIM networks based on the IDE and PCB baseline achieve the state-of-the-art results, therefore proving the effectiveness of the DIM network in ReID issue. Furthermore, because the DIM network does not extra labels, we propose a TF-DIM network which has a promising performance in unsupervised ReID issue.


%

\section*{Acknowledgment}
This work is supported by National Natural Science Foundation of China under Grant U1613214 and National Key R\&D Program of China under Grant 2017YFC0821402.

\ifCLASSOPTIONcaptionsoff
  \newpage
\fi



\bibliographystyle{IEEEtran}
\bibliography{IEEEabrv,ref.bib}
\end{document}